\documentclass[conference]{IEEEtran}
\IEEEoverridecommandlockouts
\usepackage{cite}
\usepackage{amsmath,amssymb,amsfonts}
\usepackage{algorithmic}
\usepackage{graphicx}
\usepackage{textcomp}
\usepackage{xcolor}
\usepackage{makecell}
\usepackage{array}
\usepackage{longtable}
\usepackage{caption}
\usepackage{float}
\usepackage{afterpage}
\usepackage{placeins}

\def\BibTeX{{\rm B\kern-.05em{\sc i\kern-.025em b}\kern-.08em
    T\kern-.1667em\lower.7ex\hbox{E}\kern-.125emX}}
\begin{document}

\title{Can bidirectional encoder become the ultimate winner for downstream applications of foundation models?\\
}

\author{
\IEEEauthorblockN{Lewen Yang, Xuanyu Zhou, Juao Fan, Xinyi Xie, Shengxin Zhu}
\IEEEauthorblockA{
\textit{Guangdong Provincial Key Laboratory of Interdisciplinary Research and Application for Data Science} \\
\textit{BNU-HKBU United International College}, Zhuhai, China\\
\{r130026179, r130026219, s230026033, r130026165\}@mail.uic.edu.cn, \{shengxinzhu\}@bnu.edu.cn
}
}
\maketitle

\begin{abstract}
Over the past few decades, Artificial Intelligence(AI) has progressed from the initial machine learning stage to the deep learning stage, and now to the stage of foundational models. Foundational models have the characteristics of pre-training, transfer learning, and self-supervised learning, and pre-trained models can be fine-tuned and applied to various downstream tasks. Under the framework of foundational models, models such as Bidirectional Encoder Representations from Transformers(BERT) and Generative Pre-trained Transformer(GPT) have greatly advanced the development of natural language processing(NLP), especially the emergence of many models based on BERT. BERT broke through the limitation of only using one-way methods for language modeling in pre-training by using a masked language model. It can capture bidirectional context information to predict the masked words in the sequence, this can improve the feature extraction ability of the model. This makes the model very useful for downstream tasks, especially for specialized applications. The model using the bidirectional encoder can better understand the domain knowledge and be better applied to these downstream tasks. So we hope to help understand how this technology has evolved and improved model performance in various natural language processing tasks under the background of foundational models and reveal its importance in capturing context information and improving the model's performance on downstream tasks. \par
This article analyzes one-way and bidirectional models based on GPT and BERT and compares their differences based on the purpose of the model. It also briefly analyzes BERT and the improvements of some models based on BERT.  The model's performance on the Stanford Question Answering Dataset(SQuAD) and General Language Understanding Evaluation(GLUE) was compared.
\end{abstract}
\section{Introduction}
\begin{figure*}[t]
\centerline{\includegraphics[width=\textwidth]{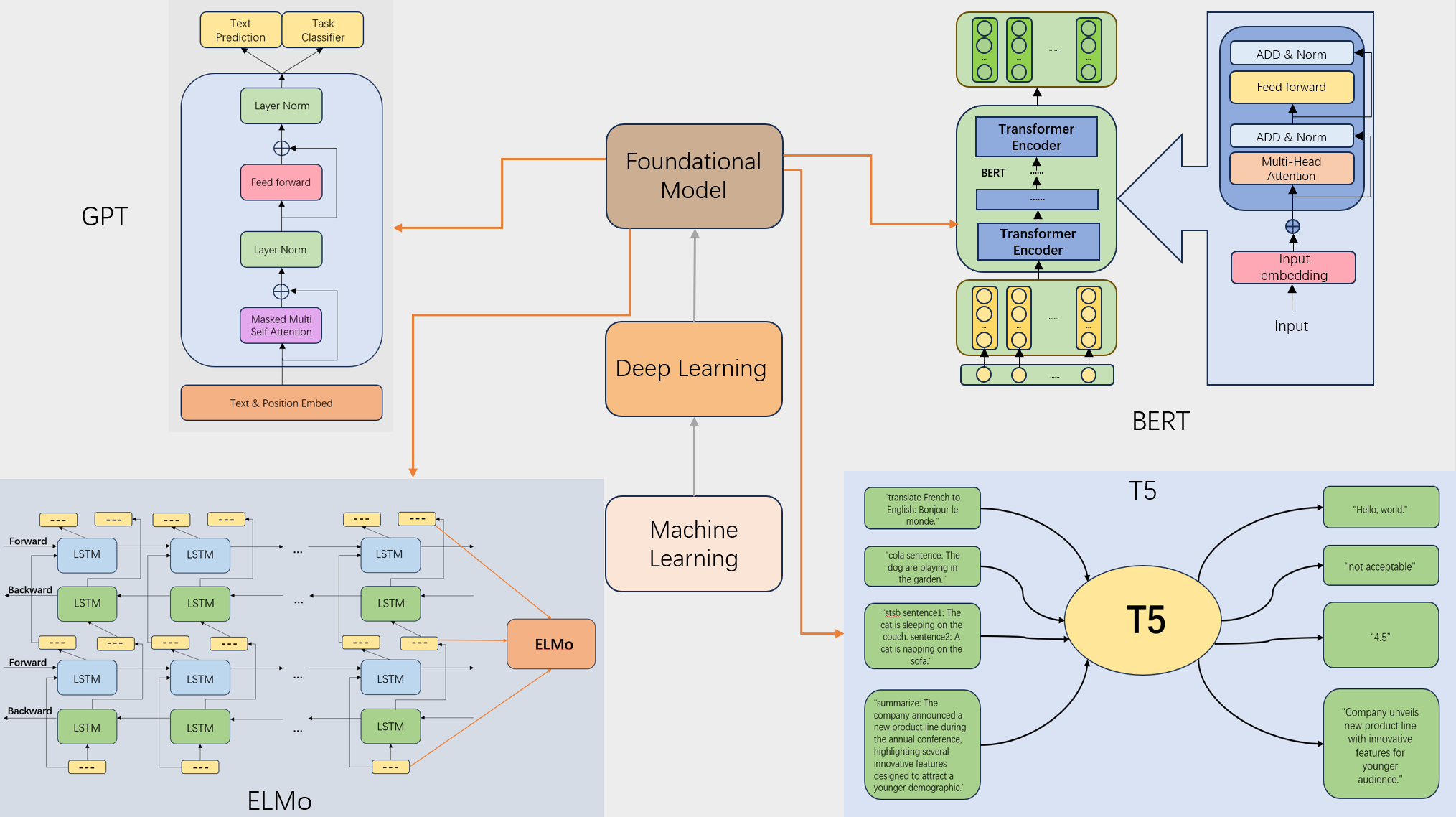}}
\caption{Foundational Model Introduction}
\label{fig:Foundational Model Introduction}
\end{figure*}

In 2021, models with high transferability and self-supervised learning were redefined as foundational models (Refer to the image in Figure \ref{fig:Foundational Model Introduction}), marking a new stage in the development of AI\cite{b27}. The popularity of GPT\cite{b29} also made more people see the potential of foundational models. The high transferability of foundational models is analogous to the smartphone, where developers create an initial yet well-architected model that is ready for immediate use, while users have the flexibility to customize it according to their specific needs. Clearly, this is a highly efficient model, where people don't need to build a house from scratch based on their needs. Moreover, self-supervised learning not only makes the model more transferable, but also greatly reduces the cost of using data. Current general large models can be viewed as foundational models, they have very extensive knowledge and can be used directly or slightly adjusted to achieve higher accuracy. There are two important branches: the one-way model led by GPT and the bidirectional model led by BERT\cite{b1}. GPT has achieved great success in practical applications in recent years and has gained a lot of traffic. However, we are currently in an era where larger models tend to perform better. Simply increasing the size of a model does not necessarily mean that it understands what it is generating. Describing GPT-like one-way generative models as having ``learned a language'' may be more appropriate\cite{b30}. Foundational one-way models generate things more like the most likely word following the previous word.\par
On the other hand, models like BERT are different. We can simply differentiate them based on their purpose. One-way models aim to predict what will come next, while bidirectional models are more about extracting features, in other words, understanding the meaning of each word. Of course, one-way models also extract features, but they always do so in a sequential order. However, real humans do not completely process information in a one-way manner. For example, ``I am happy because the weather is good.'' A one-way model cannot understand that the reason why I am happy is the weather is good, because the one-way model predicts the next word and gives the word with the highest probability. But bidirectional models can process information from both directions, allowing the model to better understand the whole sentence and context , and achieve better results.\par
Better results also make models with bidirectional encoders have a variety of applications and improvements. But as the models become larger, due to the inherent complexity of bidirectional models, they require more time and computing power than one-way models, and driven by interests, it is clear that pure one-way models have a higher cost performance at present. Of course, understanding and being able to do are certainly different. With the improvement of computing power, the situation may change in the future.
\section{The Origin of Bidirectional Encoders and BERT}
\subsection{Two sites of foundational model}
GPT and BERT can be considered among the earliest foundational models, representing two different approaches. GPT is usually called a one-way language model, also known as a causal language model, and it is a recurrent model. This type of model calculates the probability of the next word in a single direction. In contrast, BERT is considered a bidirectional language model, despite the lack of a unified definition for bidirectional language models in the field. There is a consensus that a bidirectional language model means considering both forward and backward information when processing current words, and its training goal is to obtain better representational capacity(better feature extraction ability). The birth of GPT and BERT starts with Transformer\cite{b28}, which is mainly composed of two parts: encoder and decoder. Transformer introduces a self-attention mechanism. For each position in the sequence, the self-attention mechanism computes the query, the key and value vectors, and the attention weight, and uses the attention weight to weigh the sum of the value vectors to get the final output representation. In the Transformer encoder architecture, each encoding layer's multi-head self-attention sublayer will interact with each position in the input sequence with all other positions in the sequence, so that the self-attention layer when encoding a word will consider the entire sentence's words when encoding a word. However, in the decoder, the self-attention layer uses a mask matrix so that each position in the sequence can only see the sequence before it and the positions behind it will be hidden. GPT uses the decoder part of the Transformer, while BERT uses the encoder part. Because the encoder of self-attention mechanism considers both forward and backward information, we also call it bidirectional encoder.
\subsection{BERT and its applications}
Both encoders and decoders have very strong representation capabilities thanks to the multi-head attention mechanism of Transformer, which makes it possible to train models on unlabeled datasets. The pre-training and fine-tuning paradigm then emerged in the NLP field. GPT is the earliest one, which introduced the concept of semi-supervised learning, later also known as self-supervised learning. This mainly refers to training the model on unlabeled datasets during the pre-training stage, which is unsupervised learning, allowing the model to extract more effective features\cite{b29}. Then, the model is fine-tuned using labeled datasets specific to the application scenarios, which is supervised learning, allowing the model to have better performance in the corresponding application scenario. For example, researchers can use large amounts of unlabeled data from various fields for pre-training, and then use labeled data from the medical field to fine-tune the model so that it can be better applied to medical problems, which researchers will explain in detail later. BERT, as a later model, inherited the experience of its predecessors and brought the development of the foundational model to a new stage\cite{b27}. \par
BERT builds on ideas from GPT by using the Transformer encoder and the pre-training and fine-tuning approach. Additionally, it incorporates concepts from Embeddings from Language Models (ELMo)\cite{b31}, but in a different way. model's bidirectional approach. The method used by ELMo is to pre-train word vectors using two one-way Long Short-Term Memory networks(LSTMs)\cite{b32}, thus changing the traditional word embedding representation. In traditional word embedding representations, such as Word2Vec\cite{b33} and Global Vectors for Word Representation(GloVe)\cite{b34}, each word has a fixed vector and does not consider the different meanings of words in different contexts. ELMo, on the other hand, assigns a vector to each word that varies with the context of the entire sentence by combining the hidden states of a forward LSTM and a backward LSTM. Therefore, the representation of each word is based on the current entire sentence, resulting in a dynamic word vector. This solves the problem of ambiguity of words in different sentences. Meanwhile, BERT has an excellent deeply bidirectional mechanism, unlike ELMo which only performs bidirectional operation at the surface level. BERT considers contextual information from both directions at every layer of the network, which can capture more comprehensive contextual information. Therefore, compared to sequential models such as GPT, BERT has a significant advantage in feature extraction. Excellent feature extraction ability allows the model to be deeper, which means the model will have stronger generalization ability and can solve multiple different types of problems with a single model\cite{b1}.\par
Although BERT is not the first to use pre-training and fine-tuning in NLP, it's the one that made it popular. Because of its powerful representation ability, BERT has significantly better pre-training and fine-tuning results than GPT. At the same time, BERT also marks the beginning of the model size competition, and the trend of better results with larger models can be seen from here. Later, when GPT-2\cite{b35} was developed, the research team also found that even with significant improvements to the model, the results were not much better than those of BERT, so they took a different approach which led to \textit{``Language Models are Few-Shot Learners''}(GPT-3)\cite{b36}.\par
BERT achieves such good results not only because of its bidirectional encoder, but also because of two important tasks it performs during pre-training: Mask Language Model (MLM) and Next Sentence Prediction (NSP)\cite{b1}.
\begin{itemize}
    \item \textbf{MLM: }Its working principle is to randomly mask certain words in a sentence and then train the model to predict these masked words based on the context (similar to filling in the missing parts of a sentence). This approach allows the model to learn bidirectional information by considering the context simultaneously, whereas traditional models only process text in one direction. 
    \item \textbf{NSP: }And the goal of this task is to enable the model to understand the relationships between sentences. Simply put, it achieves this by predicting whether the following sentence is the logical next sentence of the preceding one. This helps the model grasp relational information between sentences, thereby optimizing its performance. 
\end{itemize}\par
During fine-tuning, the model only needs to be trained briefly to adapt it to a specific NLP task. The main process of fine-tuning is to initialize the model with learned parameters from pre-training and further train it on labeled data of the target task. For each specific task, a simple output layer is usually added to the BERT model, which can transform the pre-trained universal language representation into the prediction results (or convert the language which is more suitable for machines into the language that is more suitable for humans and meets the requirements). During fine-tuning, the model will use the labeled data from the target task to train, and adjust all parameters and then added output layer. It allows the model to focus more attention on the required task while retaining the broad language understanding acquired during pre-training. \par
That's why BERT fine-tuning has many applications. In the professional field, for example, SciBERT \cite{b19} is trained on a large scientific text corpus and used for NLP tasks in scientific literature. ClinicalBERT \cite{b20} is pre-trained on clinical text for processing electronic health records and other medical texts. BioBERT \cite{b21} is pre-trained on biomedical text (PubMed abstracts and PMC full-text articles) for biomedical NLP tasks. BERTweet \cite{b22} is the first large-scale pre-trained language model for English tweets, pre-trained on social media text (such as Twitter) for processing informal texts. Of course, there are also fine-tuned models for various languages, such as CamemBERT \cite{b23} and FlauBERT \cite{b24} for French, BERTje \cite{b25} for Dutch, and AraBERT \cite{b26} for Arabic, etc.
\subsection{Disadvantges of Bidirectional Model}
Of course, higher representational ability and stronger understanding ability also come with a cost. Because of utilizing contextual information simultaneously, bidirectional models actually require more computation and time resources at every step of training than one-way models\cite{b3}. Additionally, due to the complex structure of bidirectional models, which handle information in both directions, it is more difficult to explain the operation of the model\cite{b38}. Furthermore, since bidirectional models do more word embedding and feature extraction, training is done using a method similar to cloze tasks, and it seems difficult to make it perform generative tasks at present. It is more suitable for classification and judgment tasks\cite{b37}.

\section{BERT and Its Evolution}
Within just a few years, the BERT model has become a foundational tool in NLP. However, as mentioned in section II.C, while the traditional BERT model performs exceptionally well in NLP tasks, it has several limitations such as high computational resource requirements, difficulty in interpretability, and unsuitability for generative tasks.

To address these limitations, researchers have proposed a series of improved models. Here, we list some models that excel in terms of computational resource optimization, interpretability, or performance in generative tasks, aiming to facilitate understanding for researchers.
\subsection{DistilBERT and TinyBERT}

A distil version of BERT (DistilBERT)\cite{b6} employs knowledge distillation to condense the original BERT model into a lighter one while retaining most of its performance. This model significantly reduces the number of parameters and computational load, thus speeding up inference and lowering memory usage, making it ideal for resource-constrained environments.

TinyBERT\cite{b39}, through distillation techniques, compresses the BERT model into a smaller version. This model maintains high accuracy while offering fast inference speed and low resource consumption, making it suitable for real-time or embedded systems.

\subsection{ALBERT}

A lite BERT (ALBERT)\cite{b5} uses parameter sharing and factorized embedding parameterization to drastically reduce the number of parameters. This approach decreases the demand for computational resources while maintaining performance comparable to BERT, and in some tasks, even surpassing the original BERT model.

\subsection{ERNIE }

Enhanced Representation through Knowledge Integration (ERNIE)\cite{b40} improves semantic understanding through combining external knowledge bases. This knowledge injection not only boosts performance in specific domain tasks but also lets the decision-making process of the model clearer and easier to interpret.

\subsection{T5}

Text-To-Text Transfer Transformer(T5)\cite{b41} transforms all kinds of NLP tasks into text-to-text format, enabling it to handle various generative tasks. It excels in translation, summarization, and question answering, demonstrating strong capabilities in generative tasks.

\subsection{SpanBERT}

SpanBERT\cite{b10} focuses on learning span-based representations, capturing relationships and entity information in sentences better than traditional BERT. It performs exceptionally well in information extraction and relation detection tasks, offering higher interpretability.

\vspace{1em} 

These models can be applied to many downstream applications, like CycleTrans\cite{b49} and EchoMamba4Rec\cite{b50} which can expands BERT application idea. For example, CycleTrans effectively integrates patients' EMR data through the combination of cross-attention mechanisms and bidirectional encoders, achieving bidirectional transmission between symptoms and medications, and attaining multiple State-Of-The-Art (SOTA) performances. Besides, EchoMamba4Rec combines bidirectional State Space Models (SSM) and frequency-domain filtering techniques. Its architecture includes bidirectional processing modules, Fast Fourier Transform (FFT), and more, enabling the model to capture complex dependencies. Following CycleTrans, EchoMamba4Rec has also achieved SOTA performances.

Additionally, we have sorted papers on Google Scholar that cited the foundational BERT model by citation count and selected some as representatives of BERT's wide application in various fields. These papers are then arranged chronologically, summarizing each model's main contributions in terms of approach and performance. This compilation aims to provide researchers with useful references. Figure 2 is the timeline of these model Details can be found in Appendix 1.

\begin{figure*}[t]
\centerline{\includegraphics[width=0.8\textwidth]{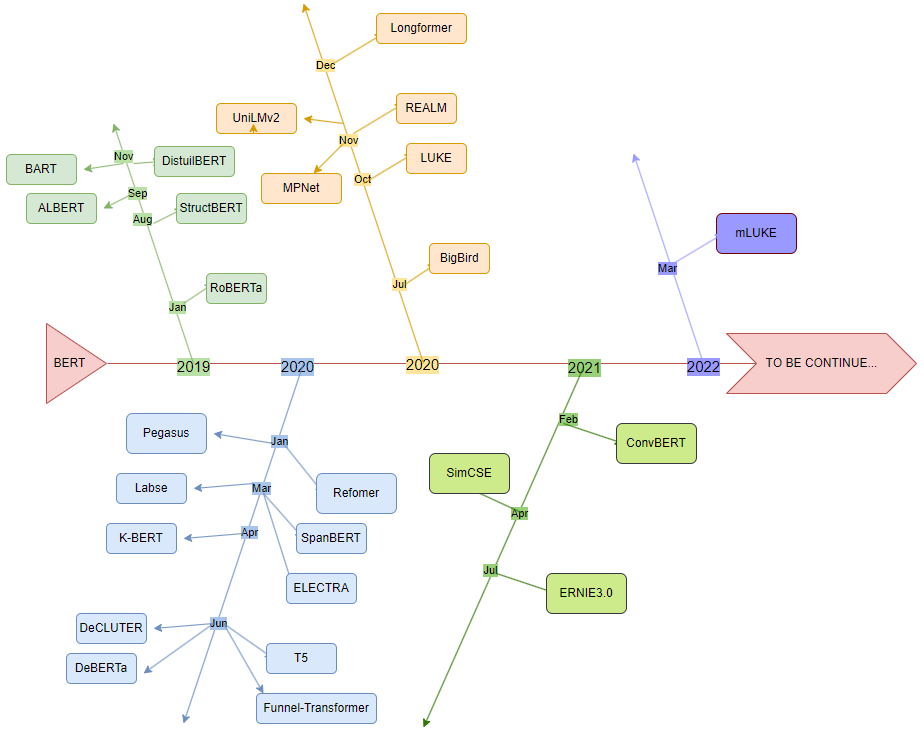}}
\caption{BERT Model Timeline}
\label{fig:BERT model timeline}
\end{figure*}

\begin{table*}[htbp]
\caption{Model Performance Comparison}
\begin{center}
\resizebox{\textwidth}{!}{
\begin{tabular}{|c|c|c|c|c|}
\hline
\textbf{Model} & \textbf{\textit{Params}} & \makecell{\textbf{\textit{GLUE}} \\ \textbf{\textit{Avg.}}} & \makecell{\textbf{\textit{SQuAD 1.1}} \\ \textbf{\textit{F1/EM}}} & \makecell{\textbf{\textit{SQuAD 2.0}} \\ \textbf{\textit{F1/EM}}} \\
\hline
BERT (Devlin et al., 2018)\cite{b1} & 340M(334M) & 82.1 & 93.2/87.4 & 83.1/80.0 \\
\hline
XLNet (Yang et al., 2019)\cite{b2} & 355M & 90.5 & \textit{*95.1}/\textit{*89.7} & 90.7/87.9 \\
\hline
RoBERTa (Liu et al., 2019)\cite{b3} & 355M & 88.5 & \textit{*94.6}/\textit{*88.9} & 89.8/86.8 \\
\hline
StructBERT (Wang et al., 2019)\cite{b4} & 340M & 83.9 & \textit{*85.2}/\textit{*92.0} & - \\
\hline
ALBERT (Lan et al., 2019)\cite{b5} & 235M & 89.4 & 95.5/90.1 & 91.4/88.9 \\
\hline
DistilBERT (SanH et al., 2020)\cite{b6} & 66M & 77 & \textit{*86.9}/\textit{*79.1} & - \\
\hline
BART (Lewis et al., 2020)\cite{b7} & 374M & - & 94.6/88.8 & 89.2/86.1 \\
\hline
ELECTRA (Clark et al., 2020)\cite{b8} & 335M & 89.5 & \textit{*94.9}/\textit{*89.7} & 91.4/88.7 \\
\hline
Funnel-Transformer (Dai et al., 2020)\cite{b9} & 488M & 89.7 & \textit{*94.7}/\textit{*89.0} & \textit{*90.4}/\textit{*87.6} \\
\hline
SpanBERT (Joshi et al., 2020)\cite{b10} & 340M & 82.8 & 94.6/88.8 & 88.7/85.7 \\
\hline
ConvBERT (Jiang et al., 2020)\cite{b11} & 106M & 86.4 & 90.0/84.7 & 83.1/80.6 \\
\hline
MPNet (Song et al., 2020)\cite{b12} & 110M & 86.5 & \textit{*92.7}/\textit{*86.9} & 85.8/82.8 \\
\hline
LUKE (Yamada et al., 2020)\cite{b13} & 483M & - & 95.4/90.2 & - \\
\hline
UNILMv2 (Bao et al., 2020)\cite{b14} & 110M & 87.3 & 92.0/85.6 & 83.6/80.9 \\
\hline
DeBERTa (He et al., 2021)\cite{b15} & 433M & 90.0 & 95.5/90.1 & 90.7/88.0 \\
\hline
\multicolumn{5}{l}{$^{\mathrm{a}}$The dev test is annotated with ``*''} \\
\multicolumn{5}{l}{$^{\mathrm{b}}$Note: Missing results in literature are signified by ``-''} \\
\end{tabular}
}
\end{center}
\label{tab:model_performance}
\end{table*}

\section{Model Performance Comparison}
Since most models are evaluated using GLUE and SQuAD, we use these two methods to compare some models[Table \ref{tab:model_performance}].
\subsection{GLUE}

Human language understanding ability is universal, flexible, and robust. In contrast, most word-level and above NLU models are designed for specific tasks and are difficult to handle out-of-domain data.\par
Wang et al.(2019) proposed the GLUE benchmark\cite{b16}, which covers a set of natural language understanding (NLU) tasks, including question answering, sentiment analysis, and semantic inference, and also provides an online platform for evaluating, comparing, and analyzing models. The GLUE framework prioritizes models that can efficiently learn from samples and effectively transfer knowledge between tasks, thus representing language knowledge in the best possible way. Since GLUE's goal is to advance general NLU systems, we designed this benchmark test to require models to share a large amount of knowledge (e.g., trained parameters) across all tasks while still retaining some task-specific components.\par
GLUE includes the following tasks:
\begin{itemize}
    \item \textbf{Corpus of Linguistic Acceptability (CoLA): }judging the grammatical correctness of sentences\cite{b42}.
    \item \textbf{Stanford Sentiment Treebank 2 (SST-2): }single-sentence sentiment classification task\cite{b43}.
    \item \textbf{Microsoft Research Paraphrase Corpus (MRP): }sentence pair semantic equivalence judgment\cite{b44}.
    \item \textbf{Semantic Textual Similarity Benchmark (STS-B): }sentence pair semantic similarity score\cite{b45}.
    \item \textbf{Quora Question Pairs (QQP): }question pair semantic equivalence judgment.
    \item \textbf{Multi-Genre Natural Language Inference (MNLI): }judging the inference relationship between sentence pairs\cite{b46}.
    \item \textbf{Question Natural Language Inference (QNLI): }inference task based on question and sentence\cite{b16}.
    \item \textbf{Recognizing Textual Entailment (RTE): }text entailment task\cite{b47}.
    \item \textbf{Winograd Natural Language Inference (WNLI): }inference task based on Winograd schema\cite{b48}.
\end{itemize}\par
The GLUE benchmark provides a diverse set of tasks, allowing researchers to comprehensively evaluate the performance of models on different types of tasks and thus gain insights into their general language understanding ability.
\subsection{SQuAD}
SQuAD tests the models' abilities in reading comprehension and information extraction, which is an important indicator of the models' understanding and generation capabilities\cite{b17}.
\paragraph{SQuAD 1.1}Known as Reading Comprehension (RC), comprehending written text and responding to questions about it,  presents a significant challenge for machines. This task demands a deep understanding of natural language and worldly knowledge. SQuAD comprises 107,785 pairs of questions and answers based on 536 articles. Systems are required to choose the answer from numerous potential spans within the passage, necessitating the ability to handle a large number of candidates. The constraint of selecting from specific spans also brings the important advantage that evaluating span-based answers is simpler than assessing freeform answers.

\paragraph{SQuAD 2.0}This challenge involves the precise formulation of various types of natural language representation and comprehension to facilitate the processing of a question and its corresponding context. Subsequently, it entails selecting an appropriate correct answer that is deemed satisfactory by humans or indicating the absence of such an answer. Each question may have a definite answer, or none at all. The unanswerable questions should appear relevant to the topic of the context paragraph. The system needs to be able to not only find the correct answer but also identify what questions are without answers\cite{b18}. Evaluation is done using F1-score and Exact Match(EM), with higher scores indicating better performance. With these features and methods, SQuAD2.0 has become an important benchmark for evaluating and improving question-answering systems.
\section{Future Work}

At present, it seems that models using bidirectional encoders have not produced a game-changing product like GPT. Many large companies have temporarily abandoned bidirectional feature extraction models and moved towards one-way generative models, such as Google switching to one-way generative models after completing T5. Of course, while bidirectional models have slowed down in recent years, their potential remains huge. Because in most cases, the understanding of each word by a one-way model is generally not as good as that of a bidirectional model, simplified, GPT-like models are aimed at generating, and they may not care about the intermediate process, because the training goal is to generate the correct answer. However, BERT-like models are different, these models aim to extract features as much as possible, that is, to understand the meaning of each word, which is obviously a more human-like way of thinking.\par
Of course, as mentioned earlier, higher computing cost and time consumption are still serious problems for bidirectional models at present, which is also the price of higher effectiveness. Only large companies have the ability to independently discover game-changing bidirectional large models. However, we believe that bidirectional models still have great potential, and we can use new acceleration inference and energy-saving training methods such as pruning, distillation, and quantization or continue researching small models to achieve this. Additionally, the recent popularity of Mamba2\cite{b62}, which has made improvements to self-attention, may also be used to speed up training time.\par
Another aspect is explainability, traditional explainability often visualizes the process of the model's output to explain the model, but this is difficult for bidirectional models. So, perhaps a way for the model to explain the process itself is more suitable for bidirectional models. Because the focus of bidirectional models is understanding, if they understand the meaning of words and sentences, then it is possible for the model to explain the origin of the answer on its own.\par
Furthermore, in today's trend towards multi-modal large models, it seems that the need for adding task-specific classification heads to BERT models to achieve better performance on specific tasks has become no longer popular. However, this does not imply that bidirectional encoders are obsolete. For instance, Google Brain introduced Unifying Language Learning Paradigms (UL2), which integrates various language model training tasks, including span masking, enabling UL2 to possess bidirectional context understanding capabilities.\cite{b61} In the future, there may also emerge improved models using bidirectional encoders.

\section{Conclusion}
The development of AI has experienced from the initial machine learning stage to the deep learning stage, and then to the foundational model stage today. The foundation model has the characteristics of pre-training, transfer learning and self-supervised learning. The pre-trained model can be applied to various downstream tasks after fine tuning. Within the realm of foundation models, models such as BERT and GPT have greatly advanced the development of natural language processing. At the same time, due to the excellent portability of BERT, the model based on BERT continue to emerge. BERT breaks through the limitation of using only one-way method for feature extraction in pre-training by using bidirectional encoder, that is, the model can capture bidirectional context information to predict mask words in the sequence, which improves the feature extraction capability of the model. The good feature extraction capability also makes subsequent improvements possible. In this paper, a brief comparison of GPT and BERT is made, and some BERT-based models and their improvements are analyzed. Finally, the performance of the bidirectional encoder-based model on SQuAD and GLUE where data can be found is collected. The models' performance in various natural language processing tasks reveals the importance of bidirectional encoders in improving the model's ability to generalize, as well as their potential for downstream applications in foundational models. 
\section*{Acknowledgment}

 This work was funded by the Natural Science Foundation of China
(12271047); Guangdong Provincial Key Laboratory of Interdisciplinary Research
and Application for Data Science, BNU-HKBU United International College (2022B1212010006); UIC research grant (R0400001-22; UICR0400008-21;
UICR0400036-21CTL; UICR04202405-21); Guangdong College Enhancement and
Innovation Program (2021ZDZX1046).

\appendix
\begin{table*}
\caption{Model Contributions and Performance}
\begin{center}
\resizebox{\textwidth}{!}{%
\begin{tabular}{|>{\centering\arraybackslash}m{2cm}|>{\centering\arraybackslash}m{1.5cm}|>{\centering\arraybackslash}m{5cm}|>{\centering\arraybackslash}m{5cm}|}
\hline
\textbf{Model} & \textbf{\textit{Time}} & \textbf{\textit{Contribution (approach)}} &  \textbf{\textit{Contribution (performance)}}\\
\hline
BERT (Devlin et al., 2018)\cite{b1} & 10/2018 & Pre-training with bidirectional Transformer for the first time & Significantly improve the performance of natural language understanding tasks \\
\hline
RoBERTa (Liu et al., 2019)\cite{b3} & 1/2019 & Proposed an optimized pre-training method with better hyperparameter tuning & Enhanced BERT's overall performance \\
\hline
DistilBERT (Sanh et al., 2019)\cite{b6} & 10/2019 & Applied knowledge distillation to create a lightweight BERT version & Reduced model size and computation \\
\hline
ALBERT (Lan et al., 2019)\cite{b5} & 9/2019 & Used parameter sharing and factorization techniques & Reduced model parameters and improved parameter efficiency \\
\hline
StructBERT (Wang et al., 2019)\cite{b4} & 8/2019 & Introduced word order and sentence order prediction tasks & Enhanced bidirectional representation capability \\
\hline
BART (Lewis et al., 2019)\cite{b7} & 10/2019 & Introduced a method combining bidirectional encoder and autoregressive decoder & Applied for generative tasks and text reconstruction \\
\hline
Reformer (Kitaev et al., 2020)\cite{b51} & 1/2020 & Used locality-sensitive hashing and reversible residual layers & Reduced computational complexity of bidirectional Transformer \\
\hline
Pegasus (Kitaev et al., 2020)\cite{b52} & 1/2020 & Designed pre-training tasks to simulate abstractive summarization & Improved performance on abstractive summarization tasks \\
\hline
SpanBERT (Joshi et al., 2020)\cite{b10} & 3/2020 & Improved masking scheme and training objectives & Better representation and prediction of text spans \\
\hline
ELECTRA (Clark et al., 2020)\cite{b8} & 3/2020 & Introduced a generator-discriminator framework & Increased pre-training efficiency \\
\hline
LaBSE (Feng et al., 2020)\cite{b53} & 3/2020 & Introduced a new pre-training and dual-encoder fine-tuning method & Achieved state-of-the-art performance in bi-text mining \\
\hline
K-BERT (Liu et al., 2020)\cite{b54} & 4/2020 & Enhanced pre-trained language models with entity-aware mechanisms[pre-train]& Improved performance on tasks involving entity information understanding[performance]\\
\hline
DeCLUTR (Giorgi et al., 2020)\cite{b55} & 6/2020 & Proposed a self-supervised sentence-level objective[architecture] & Generated generalized embeddings for sentences and paragraphs without labeled data[application]\\
\hline
T5 (Raffel et al., 2020)\cite{b41} & 6/2020 & Introduced a unified text-to-text framework for NLP tasks[architecture] & Enhanced task generalization[performance]\\
\hline
Funnel-Transformer (Dai et al., 2020)\cite{b9} & 6/2020 & Employed a funnel structure[architecture] & Reduced computational complexity while maintaining bidirectional representation capability[efficiency]\\
\hline
DeBERTa (He et al., 2020)\cite{b15} & 6/2020 & Enhanced decoding and disentangled attention mechanism[architecture] & Improved contextual information understanding[performance]\\
\hline
BigBird (Zaheer et al., 2020)\cite{b56} & 7/2020 & Utilized sparse attention mechanisms[architecture] & Enabled processing of longer text sequences[performance]\\
\hline
LUKE (Yamada et al., 2020)\cite{b13} & 10/2020 & Combined knowledge graph information with self-attention[architecture] & Enhanced performance in knowledge-intensive tasks[performance]\\
\hline
REALM (Guu et al., 2020)\cite{b57} & 11/2020 & Incorporated a latent knowledge retriever[pre-train] & Captured knowledge in a modular and interpretable way[performance]\\
\hline
MPNet (Song et al., 2020)\cite{b12} & 11/2020 & Combined masked language modeling and permuted language modeling[pre-train] & Enhanced representation capability[performance]\\
\hline
UniLMv2 (Bao et al., 2020)\cite{b14} & 11/2020 & Integrated multi-task learning and cross-task consistency[architecture] & Enhanced bidirectional Transformer representation capability[performance]\\
\hline
Longformer (Beltagy et al., 2020)\cite{b58} & 12/2020 & Combined local attention and global attention mechanisms[architecture] & Efficiently processed long texts[performance,efficiency]\\
\hline
ConvBERT (Jiang et al., 2021)\cite{b11} & 2/2021 & Combined convolutional neural networks with bidirectional Transformer[architecture] & Reduced computational complexity and improved representation capability[performance,efficiency]\\
\hline
SimCSE (Gao et al., 2021)\cite{b59} & 4/2021 & Applied simple contrastive learning[architecture] & Improved performance of bidirectional Transformer in sentence embedding tasks[performance]\\
\hline
ERNIE 3.0 (Sun et al., 2021)\cite{b40} & 7/2021 & Proposed a unified framework combining auto-regressive and auto-encoding networks[architecture] & Improved knowledge representation in multilingual settings[performance]\\
\hline
mLUKE (Ri et al., 2022)\cite{b60} & 3/2022 & Extended LUKE to multilingual versions[architecture] & Enhanced knowledge representation in multilingual environments[performance]\\
\hline

\end{tabular}
}
\end{center}
\label{tab:model_contribution}
\end{table*}
\end{document}